\documentclass[10pt, a4paper]{article}

\usepackage[final]{lrec2026} 

\usepackage{algorithmic}
\usepackage{booktabs} 
\usepackage{amsmath}
\usepackage{svg}
\usepackage{graphicx}
\usepackage{xcolor}
\usepackage{mdframed}
\usepackage{float}
\usepackage{multirow}

\title{Cross-Lingual Stability and Bias in Instruction-Tuned Language Models for Humanitarian NLP}

\name{
\\Poli Nemkova$^{1}$, Amrit Adhikari$^{1}$,
Matthew Pearson$^{2}$, Vamsi Krishna Sadu$^{1}$, Mark V. Albert$^{1}$
}

\address{
$^{1}$University of North Texas, Denton, TX, USA \\
$^{2}$Davidson College, Davidson, NC, USA \\
\texttt{poli.nemkova@unt.edu, amritadhikari@my.unt.edu, malpearson@davidson.edu,} \\
\texttt{vamsikrishnasadu@my.unt.edu, mark.albert@unt.edu}
}

\abstract{
Humanitarian organizations face a critical choice: invest in costly commercial APIs or rely on free open-weight models for multilingual human rights monitoring. While commercial systems offer reliability, open-weight alternatives lack empirical validation—especially for low-resource languages common in conflict zones.
This paper presents the first systematic comparison of commercial and open-weight large language models (LLMs) for human-rights-violation detection across seven languages, quantifying the cost-reliability trade-off facing resource-constrained organizations.
Across 78{,}000 multilingual inferences, we evaluate six models—four instruction-aligned (Claude-Sonnet-4, DeepSeek-V3, Gemini-Flash-2.0, GPT-4.1-mini) and two open-weight (LLaMA-3-8B, Mistral-7B)—using both standard classification metrics and new measures of cross-lingual reliability: Calibration Deviation (CD), Decision Bias ($\Delta$Bias), Language Robustness Score (LRS), and Language Stability Score (LSS).
.
Results show that alignment, not scale, determines stability: aligned models maintain near-invariant accuracy and balanced calibration across typologically distant and low-resource languages (e.g., Lingala, Burmese), while open-weight models exhibit significant prompt-language sensitivity and calibration drift.
These findings demonstrate that multilingual alignment enables language-agnostic reasoning and provide practical guidance for humanitarian organizations balancing budget constraints with reliability in multilingual deployment.
 \\ \newline 
 }

\begin{document}

\maketitleabstract

\section{Introduction}
Large Language Models (LLMs) are increasingly deployed in high-stakes, multilingual contexts such as conflict monitoring, crisis reporting, and human rights documentation \citep{Caballero2024On, Mikhailov2023Optimizing, Nemkova2025Do, Nemkova2025Towards, croicu2025newswirenexususingtextbased}.
Yet little is known about whether LLMs reason consistently when instructions are issued in different languages—a crucial question for cross-lingual classification tasks like human rights violation (HRV) detection, where prompt-language bias could lead to uneven assessments across regions.

This issue has direct operational significance for humanitarian organizations and UN agencies using automated content analysis systems. Commercial API services (OpenAI, Anthropic, Google) deliver strong performance but impose costly per-query fees, while open-weight alternatives (LLaMA, Mistral) remove usage costs through self-hosting but remain empirically untested in low-resource conflict languages. For decision-makers, the key question is whether open-weight models are reliable enough for deployment or whether commercial models’ stability justifies their expense in high-stakes contexts.

\begin{figure}[H]
    \centering
    \includegraphics[width=0.75\linewidth]{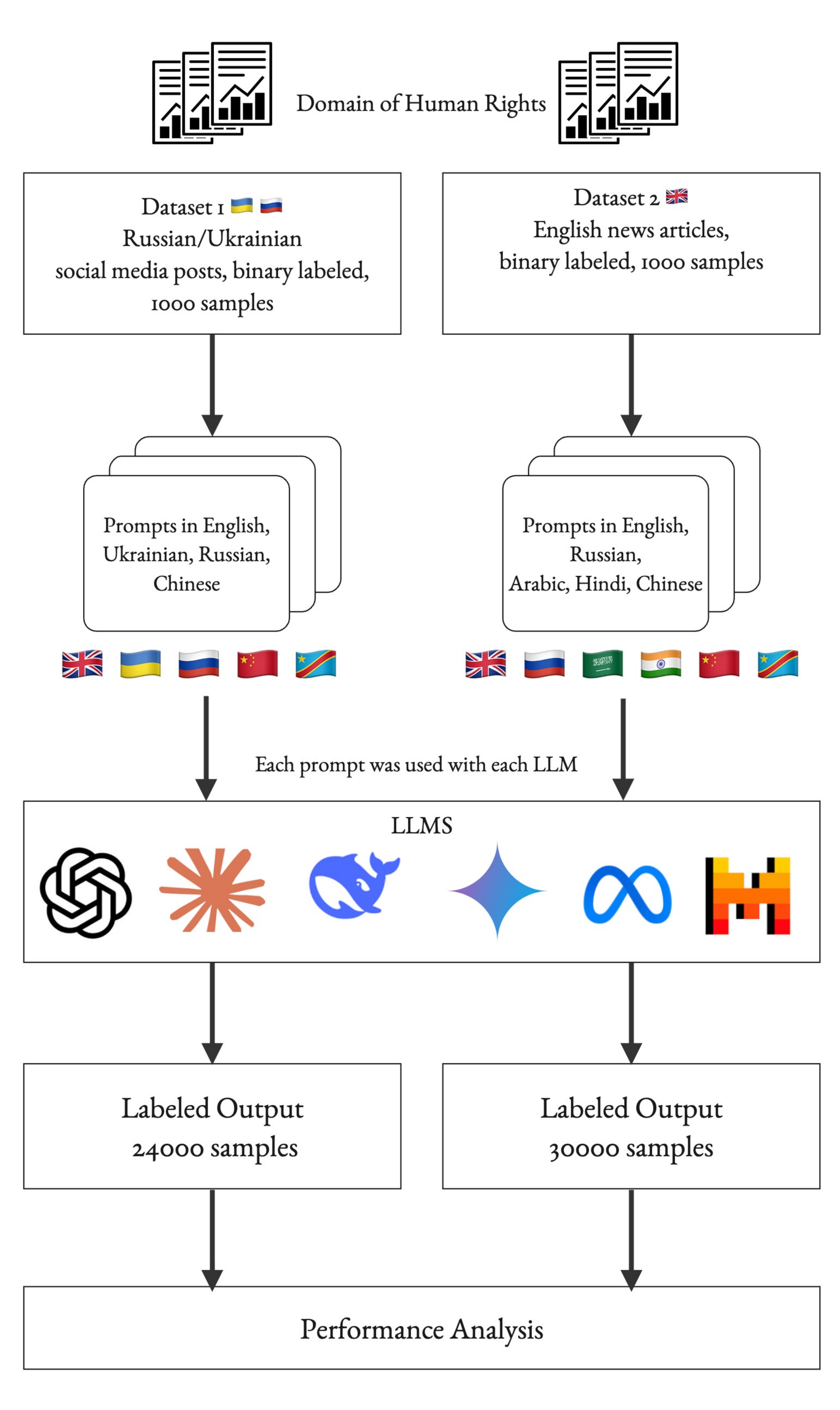}
    \caption{Experimental setup illustrating the use of two datasets, multiple multilingual prompts, and six large language models (LLMs).}
    \label{fig:sys_design}
\end{figure}

Prior multilingual evaluation studies emphasize cross-lingual transfer \citep{hong2024cross, shah2025efficient, rajaee2024analyzing} and zero-shot translation \citep{lauscher2020zero, arivazhagan2019missing}, but few examine how instruction language itself affects model reasoning.

Emerging evidence suggests that instruction tuning may induce cross-lingual consistency \citep{muennighoff2022crosslingual}, yet systematic analysis across typologically diverse prompts remains scarce—especially in socially impactful NLP domains where prompts appear in underrepresented or morphologically rich languages.

In this paper, we evaluate whether state-of-the-art LLMs exhibit prompt-language bias when classifying HRVs.
We test six models—\textit{GPT-4.1-mini}, \textit{Claude Sonnet 4}, \textit{DeepSeek-V3}, \textit{Gemini-Flash-2.0}, \textit{LLaMA-3-8B-Instruct}, and \textit{Mistral-7B}—on two datasets:
(1) a 1,000-sample Telegram HRV dataset in Russian and Ukrainian \citep{nemkova2023detecting}, and
(2) a 500-sample Human Rights Defenders (HRD) dataset in English \citep{ran2023new}.
All samples were re-labeled under multiple prompt-language conditions (English, Russian, Chinese, Arabic, Hindi, Ukrainian), while the dataset text remained unchanged.
We further include low-resource languages—Lingala\footnote{Bantu language spoken in the DRC} and Burmese\footnote{Spoken in Myanmar}—to test model reliability in high-risk linguistic regions.

Our central question is whether LLMs reason consistently across instruction languages in multilingual, high-stakes settings such as HRV detection, or whether prompt-language bias systematically alters model outputs when content remains the same.

\section{Literature Review}
\paragraph{Cultural Alignment and Value Pluralism in LLMs.}
Alignment is not monolithic: cultural value systems differ and often conflict. Recent work shows that LLMs internalize culturally specific preferences that influence downstream reasoning \citep{alkhamissi2024investigating}. To address this, researchers propose pluralistic alignment via culture-conditioned personae or agentic debates reconciling competing value targets \citep{xu2024self, ki2025multiple}. Collectively, these studies demonstrate that cultural priors shape model outputs and that single-objective alignment can obscure equity concerns in cross-cultural, high-stakes contexts.
\paragraph{Sociopolitical and Demographic Biases.}
LLMs also display ideological and demographic skews. Analyses of Persian models reveal political–economic bias across scales \citep{barkhordar2024unexpected}, while intersectional critiques expose overlapping systems of discrimination such as ableism \citep{hassan2021unpacking}. Mitigation strategies—e.g., neuron-level interventions—seek to reduce bias without severe performance loss \citep{yang2023mitigating}, yet instruction tuning can introduce new cognitive biases, highlighting trade-offs between neutrality and helpfulness.
\paragraph{Multilingual Disparities and Prompt Sensitivity.}
Performance remains uneven across languages, with large gaps in low-resource settings. Cross-Lingual-Thought (XLT) prompting and related strategies transfer reasoning from high- to low-resource languages \citep{huang2023not, qin2023cross}, while lightweight soft-prompt tuning improves transfer with minimal parameter updates \citep{philippy2024soft}. Prompt sensitivity metrics such as ProSA \citep{zhuo2024prosa} and POSIX \citep{chatterjee2024posix} further reveal that robustness varies by language, emphasizing that fairness evaluations must treat prompt formulation as a critical experimental factor.
\paragraph{Implications for Human Rights Risk Detection.}
Human rights violation (HRV) detection combines multilingual and politically charged reporting. Prior studies \citep{nemkova2025comparing, nemkova2023detecting} show that LLM performance varies by language and prompt design, while \citet{Javed2025Do} identify latent biases shaped by framing and context. Because HRV discourse is inherently politicized, alignment settings (base vs. instruction-tuned vs. culturally pluralized) influence decision thresholds and rationales \citep{alkhamissi2024investigating, xu2024self, ki2025multiple}. Evaluating whether English-authored prompts generalize—or if native-language prompts reduce false negatives—directly tests XLT-style transfer \citep{huang2023not, qin2023cross}, and varying templates estimates sensitivity bands consistent with ProSA and POSIX metrics \citep{ziems2024can, chatterjee2024posix}.
\paragraph{Bias and Subgroup Error Analysis.}
Error analysis linking model mistakes to topical or demographic dimensions (e.g., actor, defender type, or region) extends intersectional bias studies \citep{hassan2021unpacking} and political-bias work \citep{barkhordar2024unexpected}. Our work contributes an application-driven, multilingual testbed for assessing (i) alignment effects, (ii) cross-lingual prompt transfer, and (iii) prompt robustness in high-stakes HRV/HRD domains \citep{nemkova2025comparing, ran2023new}, where alignment, multilinguality, and prompt design most strongly interact.

\section{Methods}

Our objective was to examine whether Large Language Models (LLMs) produce consistent classifications of human rights violations (HRVs) when prompted in different languages, while keeping the underlying dataset text unchanged. In other words, we assess prompt language bias—whether the same instance receives a different label depending solely on the language of the instruction.

\subsection{Datasets}
We assess model behavior using two complementary datasets centered on human rights–related content: one in Russian and Ukrainian, and another in English. Both corpora include positive and negative instances of HRV mentions and were manually verified for annotation quality and balance. Details of Dataset 1 and Dataset 2 are provided in Table \label{tab:datasets}.

\begin{table*}[h!]
\centering
\small
\begin{tabular}{p{3cm} p{4cm} p{5cm} p{2cm}}
\hline
\textbf{Dataset} & \textbf{Language / Source} & \textbf{Description} & \textbf{Size} \\ 
\hline
Dataset 1: Human Rights Violation Telegram Dataset & Russian, Ukrainian (Telegram posts) & Balanced subset (500 HRV, 500 non-HRV) of posts related to the Russia–Ukraine conflict. Enables multilingual prompting, bias, and error analysis with manageable scale and cost. & 1,000  \\ 
\hline
Dataset 2: Attacks on Human Rights Defenders Reports Dataset & English (News via GDELT API) & 500 positive samples describing attacks on HR defenders + 500 negative samples retrieved and validated by three annotators for label consistency. & 1,000  \\ 
\hline
\end{tabular}
\caption{Overview of datasets used in this study.}
\label{tab:datasets}
\end{table*}

\subsection{Models}

We evaluated six state-of-the-art instruction-tuned Large Language Models (LLMs) that differ in size, architecture, and training provenance. The selection aims to cover a diverse cross-section of both proprietary and open-weight models, allowing us to examine whether language-specific biases are model-agnostic or architecture-dependent. Models overview is covered in Table 1.

\begin{table}[t]
\centering
\small
\begin{tabular}{p{2.5cm}p{4cm}}
\toprule
\textbf{Model} & \textbf{Description} \\
\midrule
GPT-4.1-mini (OpenAI, 2024) & Compact proprietary model from the GPT-4 family, optimized for efficiency and high instruction-following fidelity. \\[2pt]
Claude Sonnet~4 (Anthropic, 2024) & Mid-sized frontier model emphasizing constitutional alignment and safety-oriented fine-tuning. \\[2pt]
DeepSeek-V3 (DeepSeek AI, 2024) & Open-source model trained primarily on multilingual data, designed for improved reasoning efficiency and cross-lingual robustness. \\[2pt]
LLaMA-3-8B Instruct (Meta AI, 2024) & Open-weight transformer with 8B parameters, widely used as a research baseline for instruction-tuned multilingual capabilities. \\[2pt]
Mistral-7B (Mistral AI, 2023) & Lightweight dense model noted for strong open-source performance and efficiency on multilingual benchmarks. \\[2pt]
Gemini-Flash-2.0 (Google DeepMind, 2024) & Multimodal, instruction-tuned model variant optimized for fast inference and cross-lingual understanding. \\
\bottomrule
\end{tabular}
\caption{Overview of evaluated Large Language Models.}
\label{tab:models}
\end{table}

All models were accessed through their respective APIs or open-weight checkpoints under the same experimental protocol. Each model was queried with identical content and prompt structure, differing only in prompt language, ensuring that observed variations in predictions stem from language effects rather than task formulation or context length.

This selection reflects the practical decision space facing humanitarian organizations: four commercial models representing current paid API options, and two widely-adopted open-weight models offering cost-free self-hosted inference. Our comparison directly addresses the cost-reliability tradeoff that resource-constrained NGOs must navigate when selecting infrastructure for operational multilingual HRV detection.

\subsection{Experimental Design}

For each dataset, we re-labeled all samples using LLMs under multiple prompt-language conditions. The datasets themselves remained in their original languages, while only the prompt language varied. 

All experiments were conducted in a few-shot setting, where models received the task instruction  examples. 
We first conducted a pilot study on a small subset of samples to empirically identify the most effective base prompt for each task
The finalized English prompt was then professionally translated into the target languages and verified by native speakers to ensure semantic fidelity, linguistic naturalness, and equivalence of task framing across conditions\footnote{Translation was performed by professional translators and validated by native speakers for semantic equivalence}.

\subsection{Language Conditions}

The selection of prompt languages followed the linguistic and practical considerations described below:

\textbf{Dataset 1} (Russian/Ukrainian Telegram posts):
We used prompts in Russian and Ukrainian (matching the dataset languages), English (the dominant pretraining language for most LLMs), Chinese (representing a typologically distinct linguistic family), and a set of low-resource languages: Lingala and Burmese.

\begin{mdframed}[backgroundcolor=gray!8,rightline=false,leftline=false]
 \textbf{Base English Prompt for Dataset 1:}\\
 Instruction: You are a humanitarian state-of-the-art Human Rights Violation (HRV) detection system. Given a social media post, your job is to classify
if the post contains HRV mentions. HRV includes (but not limited to): 1) Killing civilians, 2) Destruction of civil objects, 3) Rape, torture, executions. You need to return just either "Yes" or "No" label.
Here are some examples:

\begin{itemize}
    \item Example 1. 
     Post: In Mykolaiv, after the arrival at the house, there are already five dead…\\
     Label: Yes

    \item Example 2. 
    Post: Russian aviation struck the Neptune pool in Mariupol where there were pregnant women and mothers with children under 3 years old. Orcs fighting against civilians - a special cauldron in hell is prepared for you. \\
    Label: Yes

    \item Example 3. 
    Post: The first episode is played. There will be others. And more. I will express a personal position. I cannot fail to mention this now. The Ukrainian state in its current configuration with a Nazi political regime will pose a constant, direct, and clear threat to Russia. Therefore, in addition to protecting our people and guarding the country's borders, the goal of our future actions, in my opinion, should be the full dismantling of the political regime in Ukraine.\\
    Label: No
\end{itemize}

Note that the destruction of military objects (for example, an airbase) is not considered HRV.
You need to output label only.
\\
\end{mdframed}

\textbf{Dataset 2} (English reports on human rights defenders):
We used prompts in English (the dataset’s native language), along with Arabic, Chinese, Hindi, Russian, and low-resourced Lingala and Burmese -- all representing diverse linguistic families and writing systems. 

\subsection{Implementation}

All experiments were executed in Google Colab using NVIDIA A100 GPUs. Each model was queried under all prompt-language conditions, and all outputs were recorded for subsequent quantitative and qualitative analysis.

\subsection{Evaluation Metrics}

We evaluated model performance under multilingual prompting using both conventional classification metrics and novel measures of cross-lingual robustness.

\begin{table*}[t]
\centering
\small
\begin{tabular}{ll}
\toprule
\textbf{Metric} & \textbf{Interpretation} \\
\midrule
Accuracy / Precision / Recall / F1 & Overall classification quality \\
Calibration Deviation (CD) & Symmetry of false-positive and false-negative rates across languages \\
Decision Bias ($\Delta$Bias) & Deviation from the true base positive rate (0.5) \\
Language Robustness Score (LRS) & Cross-lingual performance stability ($1 - \sigma/\mu$) \\
Language Stability Score (LSS) & Statistical invariance fraction (McNemar $p>0.05$) \\
Cohen’s $\kappa$ & Chance-adjusted agreement between prompt-language predictions \\
\bottomrule
\end{tabular}
\caption{Summary of Evaluation Metrics.}
\label{tab:metrics_summary}
\end{table*}

For each model and dataset, we computed standard binary classification metrics---\textit{accuracy, precision, recall, and F1-score}---against the human-annotated ground truth.  
These metrics quantify overall labeling quality and enable direct comparison across models and languages.

To assess \textit{linguistic robustness}, we measured intra-model consistency, defined as the proportion of samples receiving identical predictions across all prompt-language variants.  
High consistency indicates language-invariant reasoning, while large discrepancies signal potential prompt-language bias or instability in instruction-following behavior.

Beyond classical metrics, we introduced several complementary measures to capture fine-grained cross-lingual effects:  
(1) \textit{Calibration Deviation (CD),} 
CD quantifies imbalance between false-positive and false-negative rates across prompt languages.  
For model $m$ and $L$ languages:

\begin{equation}
\mathrm{CD}(m) = \frac{1}{L} \sum_{\ell=1}^{L} \left| \mathrm{FPR}_\ell - \mathrm{FNR}_\ell \right|.
\end{equation}

Lower values indicate balanced calibration across languages; higher values reflect asymmetric error behavior.

(2) \textit{Decision Bias ($\Delta$Bias)}, reflecting deviation of predicted positive rates from the ground-truth base rate;  
(3) \textit{Language Robustness Score (LRS = $1 - \sigma/\mu$)}, summarizing performance stability across languages; and  
(4) \textit{Language Stability Score (LSS)}, derived from McNemar tests comparing English and target-language predictions, indicating statistical invariance of decision patterns.  

All the described metrics are presented in Table  \ref{tab:metrics_summary}.

\section{Results and Discussion}
Across both datasets, we evaluated six large language models (LLMs) under multilingual prompting to measure the robustness of their classification behavior and the degree to which \textit{instruction-language variation} affects human-rights detection. 
Each model was prompted in six to seven languages---including four high-resource (English, Russian, Chinese, Hindi/Arabic) and two low-resource (Lingala, Burmese)---yielding more than 78,000 model-generated labels for statistical comparison.

\subsection{Cross-Lingual Performance Patterns}

All instruction-tuned frontier models (Claude~Sonnet~4, DeepSeek-V3, Gemini-Flash-2.0, GPT-4.1-mini) achieved F1-scores above 0.75 with variance $<0.03$ across languages, whereas open-weight models (LLaMA-3-8B-Instruct, Mistral-7B) fluctuated widely ($\Delta$F1 $\approx$ 0.35--0.40).  
A two-way ANOVA confirmed strong main effects of both model ($F=439.8$, $p<10^{-40}$) and prompt language ($F=13.3$, $p<10^{-14}$), as well as a significant interaction ($F=7.7$, $p<10^{-32}$), demonstrating that the \textit{impact of prompt language differs systematically by model}.  
Instruction-aligned systems showed nearly parallel performance curves across languages, while open-weights exhibited steep oscillations---evidence that multilingual alignment, rather than scale, governs cross-lingual consistency.
For humanitarian practitioners, this distinction is consequential: organizations cannot achieve commercial-model reliability simply by selecting larger open-weight alternatives. The instability observed in LLaMA-3-8B and Mistral-7B (($\Delta$F1) $ \approx $ 0.35-0.40 across languages) represents operational risk—the same violation may be detected in English but missed in Ukrainian or Burmese, creating systematic disparities in monitoring coverage across linguistic regions.


\begin{table}[t]
\centering
\small
\setlength{\tabcolsep}{3.5pt}
\begin{tabular}{lcccc}
\toprule
\textbf{Model} & \textbf{Mean F1} & \textbf{Std} & \textbf{LRS} & \textbf{Rank}\\
\midrule
Gemini-Flash-2.0 & 0.7996 & 0.0111 & \textbf{0.986} & 1\\
Claude-Sonnet-4 & 0.7758 & 0.0160 & 0.979 & 2\\
DeepSeek-V3 & 0.7726 & 0.0276 & 0.964 & 3\\
GPT-4.1-mini & 0.6871 & 0.0274 & 0.960 & 4\\
LLaMA-3-8B-Instruct & 0.6631 & 0.0545 & 0.918 & 5\\
Mistral-7B & 0.5643 & 0.1469 & 0.740 & 6\\
\bottomrule
\end{tabular}
\caption{Language-Robustness Score (LRS = $1 - \sigma/\mu$) across prompt languages.
Higher values denote greater cross-lingual stability.}
\label{tab:lrs}
\end{table}

\begin{figure*}[t]
    \centering
    \includegraphics[width=0.9\textwidth]{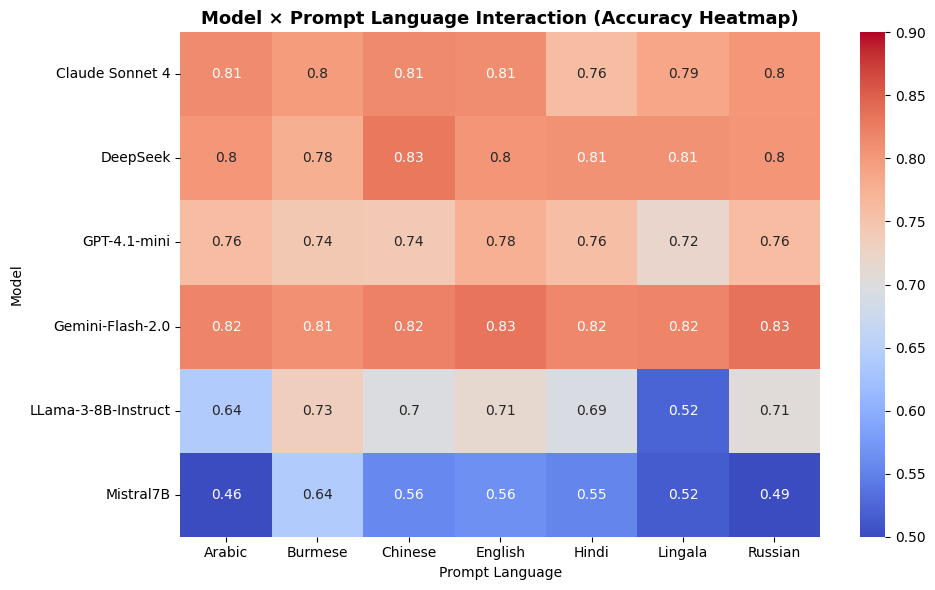}
    \caption{
    Model–Prompt Language Interaction (Accuracy Heatmap).
    Each cell shows mean classification accuracy for a given model prompted in a specific language.  
    Instruction-aligned models (Claude~Sonnet~4, DeepSeek-V3, Gemini-Flash-2.0, GPT-4.1-mini) display consistently high accuracy across languages, indicating strong cross-lingual robustness,  
    while open-weight models (LLaMA-3-8B-Instruct, Mistral-7B) exhibit pronounced prompt-language sensitivity and instability.}

    \label{fig:f1_heatmap_datasets}
\end{figure*}

\subsection{Error Structure and Calibration}

Error-taxonomy analysis revealed distinct behavioral archetypes.  
Claude, DeepSeek, and Gemini maintained balanced false-positive and false-negative rates (FPR $\approx$ 0.05--0.08; FNR $\approx$ 0.28--0.35), confirming calibrated decision thresholds.  
GPT-4.1-mini was conservative (FPR $\approx$ 0.04, FNR $\approx$ 0.45), while LLaMA and Mistral displayed erratic asymmetry.  
The derived \textit{Calibration Deviation (CD)} quantified dispersion across languages.

\begin{table}[t]
\centering
\small
\setlength{\tabcolsep}{4pt}
\begin{tabular}{lcc}
\toprule
\textbf{Model} & \textbf{CD} & \textbf{Interpretation}\\
\midrule
Gemini-Flash-2.0 & \textbf{0.021} & Stable calibration \\
GPT-4.1-mini & 0.033 & Conservative bias \\
DeepSeek-V3 & 0.064 & Mild drift \\
Claude-Sonnet-4 & 0.144 & Slight language effect \\
LLaMA-3-8B-Instruct & 0.391 & Unstable \\
Mistral-7B & 0.421 & Highly unstable \\
\bottomrule
\end{tabular}
\caption{Calibration-Deviation (CD) combining FPR/FNR variance across languages. 
Lower values indicate language-invariant calibration.}
\label{tab:calibration}
\end{table}

\subsection{Decision Bias Across Languages}

Cross-lingual bias analysis compared each model’s predicted positive rate with the true base rate (0.5).  
Frontier models exhibited mild under-prediction ($\Delta$Bias $\approx -0.12 \pm 0.04$), while LLaMA and Mistral oscillated dramatically between under- and over-prediction across languages.  
Gemini’s predictions were closest to neutral bias ($-0.11 \pm 0.01$), indicating consistent priors across languages.

\subsection{Statistical Consistency (McNemar Tests)}

Pairwise McNemar comparisons (English vs.\ target language) confirmed that Gemini-Flash-2.0 and GPT-4.1-mini retained statistically indistinguishable predictions for most languages ($p>0.05$ in four of six cases).  
Claude and DeepSeek showed small but consistent differences ($p<0.05$ for $>$80\% of languages), while open-weight models exhibited extreme divergence ($p<10^{-33}$).


\subsection{Low-Resource Languages (Lingala, Burmese)}
Despite expected degradation, aligned models maintained strong performance in Lingala and Burmese:
Claude~Sonnet~4 (F1 = 0.83), Gemini-Flash-2.0 (F1 = 0.80), DeepSeek-V3 (F1 = 0.78).  
LLaMA and Mistral dropped below 0.60~F1 and displayed inflated FPR ($>$0.8), confirming poor transfer.  
Aligned models preserved recall above 0.85, indicating effective latent transfer even for typologically distant languages.

\begin{table}[t]
\centering
\small
\setlength{\tabcolsep}{4pt}
\begin{tabular}{lcc}
\toprule
\textbf{Model} & \textbf{LSS} & \textbf{Significant Drift (\%)}\\
\midrule
Gemini-Flash-2.0 & \textbf{0.67} & 33\\
GPT-4.1-mini & 0.67 & 33\\
Claude-Sonnet-4 & 0.17 & 83\\
LLaMA-3-8B-Instruct & 0.17 & 83\\
DeepSeek-V3 & 0.00 & 100\\
Mistral-7B & 0.00 & 100\\
\bottomrule
\end{tabular}
\caption{Language Stability Score (LSS = $1 - \#(p<0.05)/N$). 
Higher scores indicate consistent predictions across prompt languages.}
\label{tab:mcnemar}
\end{table}

\section{Discussion}
The results across 78,000 multilingual inferences consistently demonstrate that \textit{alignment}, rather than model scale, is the primary factor determining cross-lingual robustness.  
Models that underwent explicit instruction alignment, such as Gemini-Flash-2.0, Claude-Sonnet-4, DeepSeek-V3, and GPT-4.1-mini, exhibited highly stable behavior across languages, maintaining consistent predictions and minimal bias drift.  
This suggests that multilingual instruction tuning creates shared latent semantics that allow these systems to reason effectively even when prompts are issued in typologically distant languages.

Although statistical tests such as ANOVA and McNemar reveal small but significant prompt-language effects, the practical impact of these differences is negligible.  
Instruction-aligned models behave in a largely language-agnostic manner, providing consistent classification outcomes across linguistic contexts—a critical property for tasks like human-rights monitoring, where fairness and reliability are paramount.  

Calibration analysis further supports this conclusion.  
Frontier models maintain balanced false-positive and false-negative rates across languages, indicating that their decision thresholds remain stable regardless of the prompt language.  
This calibration symmetry is not only a sign of model reliability but also a key fairness attribute, as it minimizes the risk of systematic under- or over-detection in specific linguistic regions.  

Performance in low-resource languages reinforces these trends.  
Models such as Claude and Gemini maintained strong F1-scores and high recall in Lingala and Burmese, demonstrating that multilingual alignment—rather than tokenizer coverage or sheer scale—drives successful transfer.  
By contrast, unaligned open-weight models like LLaMA-3-8B and Mistral-7B showed substantial volatility and prompt-language sensitivity, underscoring the fragility of models that lack explicit cross-lingual alignment.

Taken together, these findings highlight several practical implications for multilingual humanitarian AI pipelines.  
First, alignment quality outweighs model size in determining robustness; second, instruction-tuned models can be deployed reliably across diverse languages; and third, open-weight models should be paired with human verification or hybrid approaches when used in operational contexts.  
Evaluation frameworks should also integrate robustness diagnostics—such as the Language Robustness Score (LRS) and McNemar-based stability testing—to ensure cross-lingual consistency before real-world deployment.

In summary, prompt-language bias is statistically detectable yet operationally negligible in aligned LLMs.  
Gemini-Flash-2.0 achieves the highest degree of cross-lingual stability (LRS~$\approx$~0.99), whereas Mistral-7B exemplifies the instability of unaligned models.  
Overall, the results affirm that alignment—not dataset language or scale—underpins true multilingual generalization in contemporary large language models.

\subsection{Cross-Lingual Agreement Analysis}

\begin{table*}[h]
\small
\centering
\begin{tabular}{lcccc}
\toprule
& \multicolumn{2}{c}{\textbf{Dataset 1}} & \multicolumn{2}{c}{\textbf{Dataset 2}} \\
\cmidrule(lr){2-3} \cmidrule(lr){4-5}
& Correct & Error & Correct & Error \\
\midrule
Open-source    & 53 (5.3\%)  & 11 (1.1\%) & 29 (2.9\%)  & 12 (1.2\%) \\
Proprietary    & 343 (34.3\%) & 10 (1.0\%) & 376 (37.6\%) & 29 (2.9\%) \\
\midrule
\textbf{Ratio} & \textbf{34:1} & & \textbf{13:1} & \\
               & \textbf{5:1}  & & \textbf{2:1}  & \\
\bottomrule
\end{tabular}
\caption{Cross-lingual unanimous agreement. "Correct" = all models 
in category agreed on correct label; "Error" = all models agreed on 
same incorrect label. Ratios show correct:error agreement rates.}
\label{tab:agreement}
\end{table*}

To assess whether models interpret content consistently across 
languages, we identified instances where all models within each 
category reached unanimous predictions across all seven prompt 
languages—both unanimous correct classifications and unanimous 
errors (where all models made the same mistake).

Open-source models achieved unanimous correct labels on only 53 
instances (5.3\%) in Dataset 1 and 29 instances (2.9\%) in Dataset 2, 
while proprietary models reached unanimous correct classifications 
on 343 (34.3\%) and 376 (37.6\%) instances respectively—a roughly 
10× difference. Unanimous errors—cases where all models within a 
category incorrectly classified the same instances—occurred in 11 
(1.1\%) and 12 (1.2\%) cases for open-source models versus 10 (1.0\%) 
and 29 (2.9\%) for proprietary models.

The critical distinction lies in the correct-to-error ratio: 
proprietary models achieve unanimous correct classifications 34× 
more frequently than unanimous errors (343:10 in Dataset 1), while 
open-source models show only a 5:1 ratio (53:11). This demonstrates 
that proprietary systems maintain stable, accurate interpretation 
across languages, whereas open-source models exhibit low consensus 
even on straightforward cases.

\textbf{Unanimous correct instances} featured explicit harm 
indicators that transcended linguistic framing: \textit{"New footage 
of Russian soldiers' atrocities in Bucha—entire family tortured and 
burned"} or \textit{"Russian forces shelled residential Kharkiv—4 
dead, 30 injured."} In Dataset 2, overt repression like \textit{"Hong 
Kong police arrest five speech therapists over children's books 
deemed seditious"} achieved consistent cross-lingual recognition.

\textbf{Unanimous errors}—where all models within a category 
misclassified the same instances across all languages—reveal 
inherently difficult cases:

\begin{enumerate}
\item \textbf{Sarcastic/propagandistic framing}: \textit{"Fake: 
nitric-acid explosion in Luhansk—Truth: Ukrainians surrendering 
Rubizhne"}\footnote{Translated from Russian.} uses conflict 
keywords without describing actual violations.

\item \textbf{Implicit harm references}: \textit{"Missile strike 
also hit Zaporizhzhia region"} lacks explicit civilian casualty 
markers, causing systematic underprediction.

\item \textbf{Analytical/procedural discourse}: Policy commentary 
(e.g., \textit{"Fixing Nigeria Before the Fall"}) or institutional 
reports (HRW's Egypt UPR review) discuss violations abstractly 
rather than as concrete events.
\end{enumerate}

These shared error types represent genuinely ambiguous edge cases 
where even aligned models struggle. However, proprietary models' 
34:1 correct-to-error consensus ratio versus open-source models' 
5:1 ratio demonstrates that alignment enables robust cross-lingual 
agreement on the vast majority of content, while open-source models 
achieve consensus primarily on trivial cases. For humanitarian 
organizations, this gap translates directly to operational reliability: 
proprietary systems provide predictable, language-invariant 
classification, whereas open-source models' low agreement rates 
indicate inconsistent regional coverage.
\\
\subsection{Implications for Resource-Constrained Deployment}

Our findings directly inform deployment decisions for humanitarian organizations operating under budget constraints. While open-weight models eliminate per-query API costs, their substantial cross-lingual instability introduces operational risks that may outweigh cost savings in high-stakes contexts.
\textit{Cost-Reliability Tradeoffs.} Commercial models maintain consistent performance (F1 variance < 0.03) across languages, while open-weight models fluctuate dramatically (F1 variance > 0.15). For organizations monitoring violations across multiple linguistic regions, this instability creates disparate detection rates—potentially missing violations reported in certain languages while detecting equivalent content in others.
Low-Resource Language Performance. The performance gap widens critically in low-resource languages. In Lingala and Burmese—spoken in regions with active humanitarian concerns—commercial models maintain F1 > 0.78 while open-weight models drop below 0.60 with inflated false-positive rates (> 0.8). Organizations working in conflict zones with low-resource languages cannot reliably substitute open-weight alternatives.
\textit{Deployment Recommendations.} Based on our empirical results:

\begin{itemize}
    \item Multilingual monitoring requiring cross-lingual consistency: Commercial aligned models necessary
    \item Single-language or English-dominant workflows: Open-weight models may suffice with human verification
    \item Low-resource language contexts: Commercial models strongly recommended
    \item Hybrid approaches: Consider open-weight for initial filtering, commercial for final classification
    \item Infrastructure costs: Self-hosting requires GPU infrastructure, technical maintenance, and expertise that may offset API cost savings
\end{itemize}

Organizations should prioritize alignment quality and cross-lingual stability over cost minimization when operational reliability directly impacts humanitarian response.


\section{Conclusion and Future Work}

This study provides the first systematic evaluation of prompt-language robustness in large language models for human-rights monitoring.
Across 78 000 multilingual inferences, instruction-aligned models (Gemini-Flash-2.0, Claude-Sonnet-4, DeepSeek-V3) exhibit near language-invariant reasoning, while open-weight models (LLaMA-3-8B, Mistral-7B) show strong prompt-language sensitivity.
Our metrics—Calibration Deviation, Decision Bias, Language Robustness (LRS), and Language Stability (LSS)—show that multilingual alignment, not scale, governs cross-lingual reliability.
Aligned models maintain balanced calibration even in low-resource languages such as Lingala and Burmese, underscoring that alignment quality should outweigh parameter count.
For humanitarian deployment, open-weight systems are not yet dependable substitutes for commercial APIs; organizations should prioritize verified alignment and stability over cost.
Future work will extend this framework to multi-label and hybrid deployments, expand coverage of African and Indigenous languages, and track evolving cost-reliability trade-offs as open-weight models mature.

\section*{Limitations}
This study is limited to binary human-rights classification tasks and may not fully capture the complexity of multilingual reasoning required for richer contextual or multi-label analyses.  
While translation quality was manually checked, prompt phrasing differences across languages may still introduce subtle lexical or cultural bias.  
The language coverage, though diverse, remains incomplete—particularly for morphologically rich or under-represented African and Indigenous languages.  
Finally, model outputs were evaluated on fixed test sets, so potential effects of domain drift or time-evolving data are not yet quantified.

\section*{Ethics Statement}
This work contributes to research on multilingual detection of human rights violations using large language models (LLMs).  
All datasets employed are either publicly available or manually curated from open social media and news sources.  
No private, personal, or non-consensual data were collected or accessed.  

The findings are intended solely for research and humanitarian awareness purposes, not for automated surveillance or enforcement.  
We emphasize the importance of responsible and transparent use of these methods, particularly in contexts involving vulnerable communities or ongoing conflicts.  
All model evaluations were conducted in accordance with institutional ethical standards, and outputs were analyzed only in aggregate form to prevent potential harm or re-identification.  

An LLM-based assistant was used exclusively for language and style polishing; all conceptual and analytical content was authored by the researchers.


\bibliographystyle{lrec2026-natbib}
\bibliography{lrec2026-example}

\end{document}